\documentclass[letterpaper, 10 pt, conference]{ieeeconf}  

\IEEEoverridecommandlockouts                              
\usepackage{graphicx}
\usepackage{multirow}
\usepackage{cite}
\usepackage{subfigure}
\usepackage{here}

\title{\LARGE \bf
Learning Based Industrial Bin-picking\\ Trained with Approximate Physics Simulator
}

\author{Ryo Matsumura$^{1}$, Kensuke Harada$^{1,2}$, Yukiyasu Domae$^{2}$ and Weiwei Wan$^{1,2}$ 
\thanks{$^{1}$Graduate School of Engineering Science, Osaka University, 
        1-3 Machikaneyama, Toyonaka 560-8531, Japan
        {\tt\small matsumura@hlab.sys.es.osaka-u.ac.jp, \{harada, wan \}@sys.es.osaka-u.ac.jp}}%
\thanks{$^{2}$Intelligent Systems Research Institute, National Institute of Advanced Industrial Science and Technology, 
        1-1-1 Umezono, Tsukuba 305-8568, Japan
        {\tt\small yukiyasu.domae@aist.go.jp}}%
}

\begin{document}

\maketitle
\thispagestyle{empty}
\pagestyle{empty}

\begin{abstract}
In this research, we tackle the problem of picking an object from randomly stacked pile. Since complex physical phenomena of contact among objects and fingers makes it difficult to perform the bin-picking with high success rate, we consider introducing a learning based approach. For the purpose of collecting enough number of training data within a reasonable period of time, we introduce a physics simulator where approximation is used for collision checking. In this paper, we first formulate the learning based robotic bin-picking by using CNN (Convolutional Neural Network). We also obtain the optimum grasping posture of parallel jaw gripper by using CNN. Finally, we show that the effect of approximation introduced in collision checking is relaxed if we use exact 3D model to generate the depth image of the pile as an input to CNN. 
\end{abstract}

\section{Introduction}

Randomized bin-picking refers to the problem of automatically picking 
an object from randomly stacked pile. If randomized bin-picking is introduced 
to a production process, we do not need any parts-feeding machines or human workers 
to once arrange the objects to be picked by a robot. Although a number of researches 
have been done on randomized bin-picking such as 
\cite{turkey,Kristensen,Frydental,Hujazi,Ghita,Kirkgaard,Fuchs,Zuo,Domae2014,Dupuis_08,icra13,SII}, 
randomized bin-picking is still difficult due to the complex physical phenomena of contact 
among objects and fingers. 
To cope with this problem, learning based approach has been researched 
by some researchers such as\cite{case2016,Levine}. 
By using the learning based approach, it is expected that the complex physical phenomena can 
automatically be learned and that we can be realized the robotic bin-picking with high success rate. 

In this paper, we research a learning based approach for robotic bin-picking. 
We introduce CNN (Convolutional Neural Network) to predict whether or not a robot can successfully 
pick an object from the pile for given depth image of the pile and grasping pose 
of a parallel jaw gripper. 
Since our CNN outputs the success rate of picking, we search for the grasping pose maximizing the success rate. 
However, learning based bin-picking trained with CNN usually requires extremely large number 
of training data. To cope with this problem, this research aims to effectivelly collect enough number 
of training data by introducing a physics simulator. 
Here, physics simulation on randomly stacked objects with complex shape usually takes longer time than 
the physics simulation of simple shaped objects. 
For the purpose of shortening the calculation time of physics simulation used to collect the training data, 
we consider approximating the shape of objects. This approximation is applied for checking collision among 
objects and fingers. Here, although we introduce approximation in physics simulation, 
we do not want to reduce the accuracy of prediction made by CNN. 
One of the goals of our research is to give an answer to the question: 
{\it how we can relax the effect of object shape approximation on the accuracy of prediction. }

The feature of our physics simulator is that, while the shape approximation is introduced for checking collision, 
simulated depth image of the pile used as an input to CNN is obtained by using objects with the original shape. 
To check the effect of approximation on the accuracy of prediction, we consider focusing on some cases included 
in the training data where a robot successfully picked an object with approximated shape while a robot may fail 
in picking the same object with original shape. 
Our finding in the research is that {\it 
even if we approximate the object shape 
in collision checking, the effect of approximation can be relaxed if we use original shaped object to construct 
simulated depth image of the pile as an input to CNN. }

The rest of this paper is organized as follows: After introducing previous works in Section 2, 
we show the overview of our physics simulator in Section 3. In Section 4, we explain our learning based 
bin-picking method. In Sections 5 and 6, we show results by using our learning based method. 

\begin{figure}[t]
\centering {
\includegraphics[width=75mm]{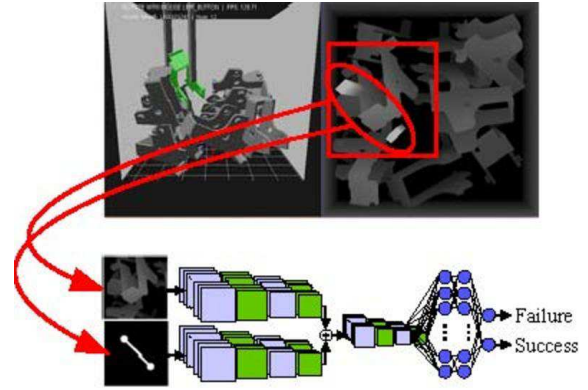}
}
\caption{Overview of learning based randomized bin-picking trained with physics simulator}
\label{fig:intro}
\end{figure}

\section{Related Works}

So far, research on industrial bin-picking has been mainly done on 
image segmentation \cite{turkey,Kristensen,Frydental,Hujazi}, 
pose identification \cite{Ghita,Kirkgaard,Fuchs,Zuo}, 
and picking method \cite{Domae2014,Dupuis_08,icra13,SII}. 

As for the research on bin-picking method, 
Ghita and Whalan \cite{Ghita} proposed to pick the top most object of the pile. 
Domae et al. \cite{Domae2014} proposed a method for determining 
the grasping pose of an object directly from the depth image of the pile. 
Some researchers such as \cite{Fuchs,Dupuis_08,icra13,SII} 
proposed methods for identifying the poses of multiple objects of the 
pile and picking one of them by using a grasp planning method. 
However, in conventional bin-picking methods, we have to carefully set up 
several parameters used in both visual recognition and grasp planning 
corresponding to each object to be picked. Since a robot usually has to pick 
a lot of objects to assemble a product, it is not easy for setting up parameters 
for all objects to be picked. 

On the other hand, learning based approaches on randomized bin-picking is 
expected to break this barrier existing in the conventional randomized bin-picking 
\cite{Levine,case2016,graspgan,Dexnet2}. 
Levine et al. \cite{Levine} proposed an end-to-end approach by using deep neural network whose 
input is a 2D RGB image. 
However, they need extremely large number of training data which was collected 800,000 times of picking trials 
for two months by using 2D RGB image of the pile. 
Recently, there are some trials on reducing the effort to collect a number of training data by using 
a method so called GraspGAN \cite{graspgan} and cloud database \cite{Dexnet2}. 
On the other hand, this research aims to collect enough number of training data within reasonable time 
by introducing an approximate physics simulation. Our method searches for the grasping posture with maximum success 
rate of picking. 


The learning approach has also been used for grasping a novel daily object 
placed on a table\cite{Curtis,Lenz,Pas,Ekvall} and for warehouse automation \cite{Zeng,RefineNet}. 
Pas et al. \cite{Pas} developed a method for 
learning an antipodal grasp of a novel object by using the SVM (Support Vector Machine). 
Lenz et al. \cite{Lenz} used deep learning to detect the appropriate 
grasping pose of an object. 
Zeng et al. \cite{Zeng} proposed a learning based picking method used for warehouse automation. 
However, industrial bin-picking is different from the warehouse application since 
the grasped object is not existing in our daily life and it is impossible to use the generalized object recognition methods. 

\section {Physics Simulator}\label{sec:simulator}

\begin{figure}[htb]
\centering {
\includegraphics[width=60mm]{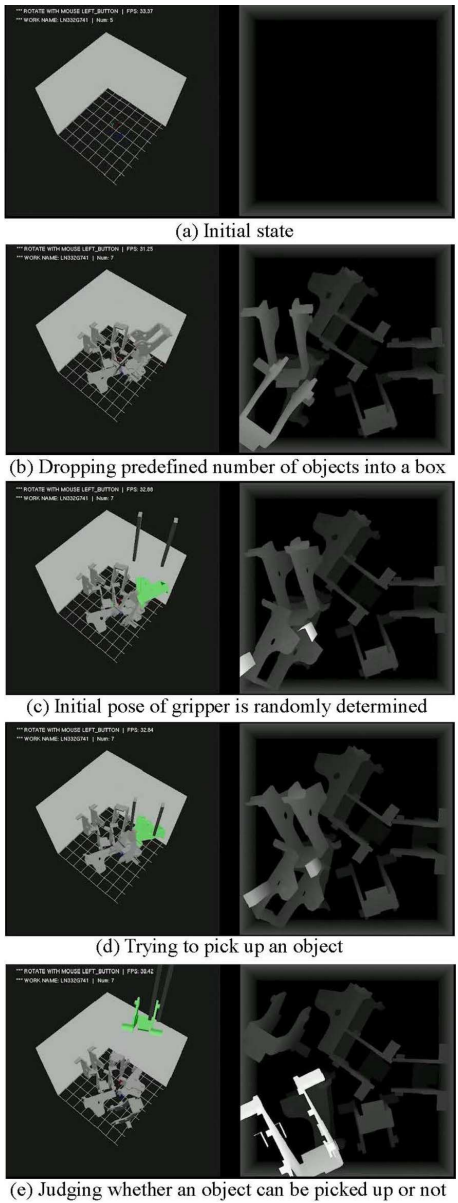}
}
\caption{Physics simulator used in this work where the left side shows the overview of simulation while the right side shows the depth image}
\label{fig:simulator}
\end{figure}

In this section, we show an overview of the physics simulator used in this research. 
We use PhysX as a physics engine. The overview of the simulator is shown in Fig. \ref{fig:simulator}. 
In the simulation world, we assume a tray where its bottom surface is 
horizontally flat. We also assume the gravity acceleration acting in a vertically downward 
direction. We use two rectangular shaped objects simulating a two-fingered parallel jaw gripper 
where a gripper can translate, rotate about the vertical axis and open/close the fingers. 

To collect training data, we first consider dropping predefined number of objects 
from predefined height with randomly defined poses. Then, we consider obtaining a 
simulated depth image of the pile assuming that a simulated 3D depth sensor is facing 
vertically downward direction. We furthermore define the gripper's horizontal position 
and orientation about the vertical axis for the gripper to grasp the top most object. 
To pick an object, the gripper first moves in the vertically downward direction, 
closes the fingers, and then moves in the vertically upward direction.
After the gripper moves up, we judge whether or not an object 
is successfully picked by checking the vertical position of objects. 
At each picking trial, we collect the following three information: 1) a depth image of the pile, 
2) gripper's horizontal position and orientation about the vertical axis, 
and 3) success/failure of picking. 

\begin{figure}[htb]
\centering {
\includegraphics[width=70mm]{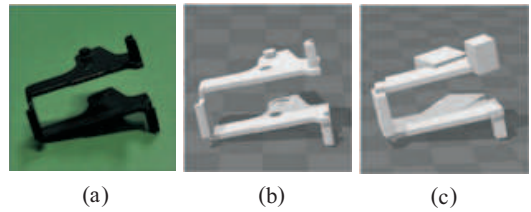}
}
\caption{Objects used in a physics simulation where (a) Real object, (b) 3D model used as input depth images of CNN, 
(c) Approximate model used for interference calculation in simulation.}
\label{work}
\end{figure}

As explained in the introduction, 
physics simulation of randomly stacked objects with complex shape usually takes longer time 
than the simulation of simple shaped objects since 
the calculation time of physics simulation usually depends on the number of contact 
points included in the simulation world. For the purpose of shortening the 
calculation time of physics simulation used to collect training data, 
we consider approximating the shape of an object. 
This approximation is used just for checking collision among objects and fingers. 
In our method, the shape of an object is approximated by a set of shape primitives such as rectangular. 
Fig. \ref{work} shows our method for approximating an object shape. 
For a given polygonal model of an object, we consider applying the convex decomposition \cite{convex} where 
it is decomposed into a set of convex shaped polygons. 
Then, for each convex shaped polygon, we consider fitting rectangular. 

\begin{figure}[htb]
\centering {
\includegraphics[width=70mm]{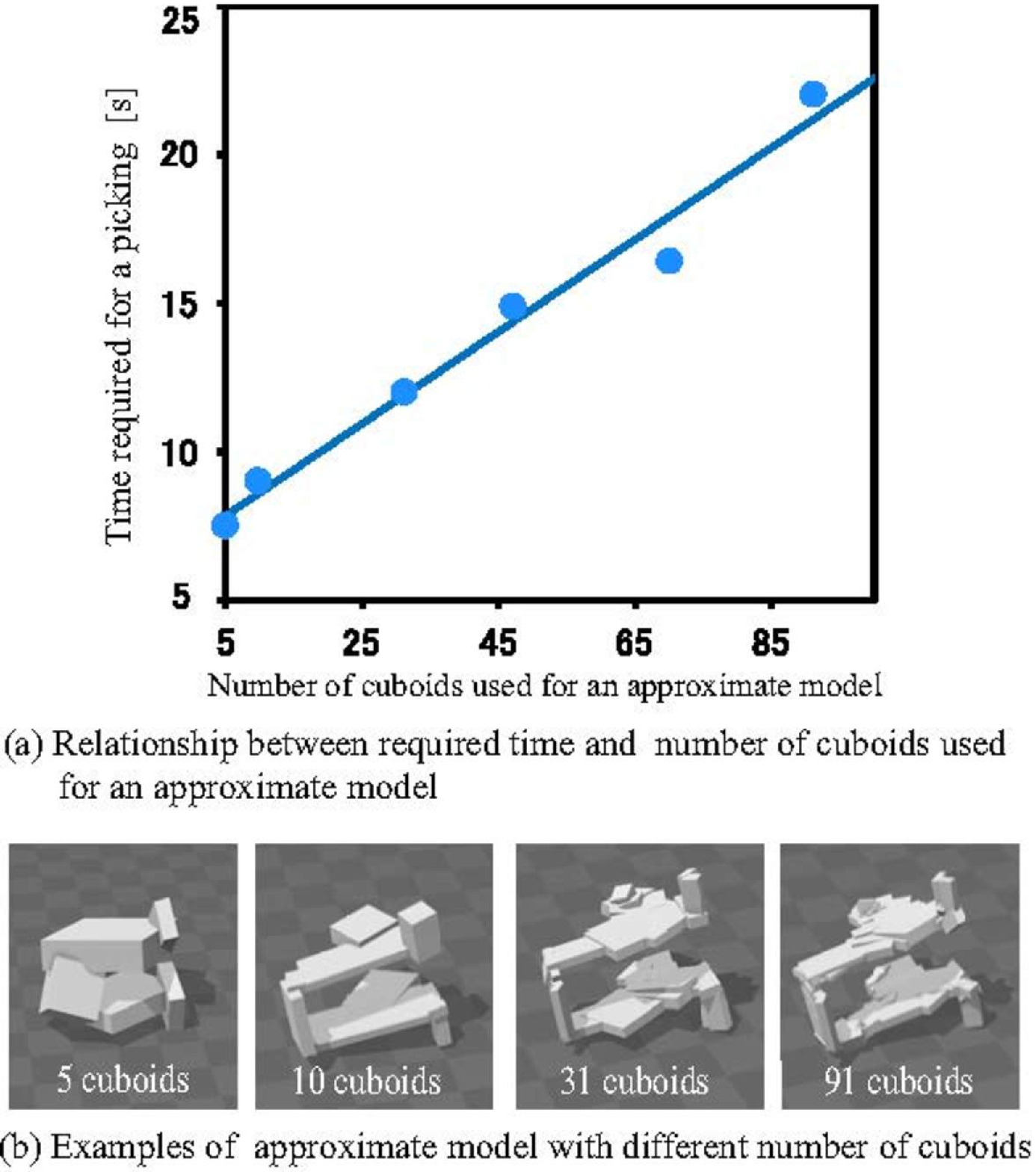}
}
\caption{Trade-off between model accuracy and time required for picking}
\label{calctime}
\end{figure}

We checked the calculation time of physics simulation as shown in Fig. \ref{calctime}. 
We performed simulation of picking an object from the pile for four times where 
each simulation includes same number of same objects with different resolution of convex decomposition. 
As shown in the figure, as the number of convex objects generated by the convex decomposition increases, calculation time 
of physics simulation also increases. In the following, we set each object 
decomposed into 10 rectangular polygons as shown in Fig. \ref{work}. 

Here, we note that, although the convex decomposition is introduced just for checking collision of physics simulation, 
it is not used to obtain a simulated depth image of the pile 
which is an input to the CNN explained in the next section. 

\section{Learning Based Approach}

This section explains our learning based approach for randomized bin-picking introduced in this research. 

\subsection{Convolutional Neural Network}\label{CNN}

\begin{figure}[thb]
\centering {
\includegraphics[width=90mm]{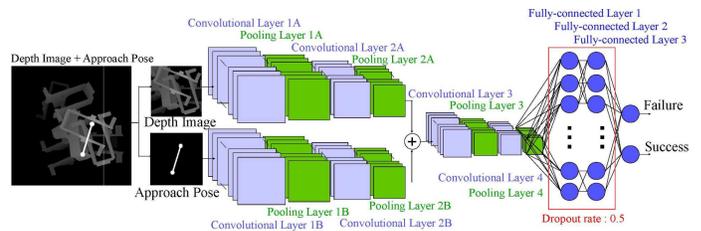}
}
\caption{Proposed architecture of CNN}
\label{cnn_figure}
\end{figure}

We use CNN (Convolutional Neural Network) \cite{grauman2011visual,russakovsky2015imagenet} 
to predict whether or not a robot can successfully pick an object from the pile. 
The overview of our CNN is shown in Fig. \ref{cnn_figure} and Table \ref{cnn_table}. 
We use a depth image of the pile (500 $\times$ 500 [pixel]) and gripper's pose before picking an object. 
Since we use a parallel jaw gripper grasping an object with upright posture, a gripper can be expressed 
by using a segment where two fingers are located at the edge. 
To reduce the time needed to train the CNN, we consider extracting 250$\times$250[pixel] subset of the pile's image. 
This is an input to the main channel of the CNN. 
On the other hand, 250$\times$250 [pixel] image of the segment expressing a gripper's pose is an input 
to the side channel of the CNN. 
Our CNN is composed of serially connected convolutional and pooling layers. 
In the pooling layer, we applied the max pooling of 2$\times$2 [pixel]. 
At the end of the convolutional and pooling layers, fully-connected layers is attached. 
The last layer of the fully-connected ones classifies success/ failure of picking. 
Success and failure rates are denoted respectively by $y_0$ and $y_1=1-y_0$ by using the following softmax function: 

\begin{eqnarray}
\centering {
y_k=\frac{{e}^{a_k}}{\sum_{i=1}^{n}{e}^{a_i}}\label{soft-max}
}
\end{eqnarray}

\noindent
where $a_k$ denotes weight of the input to the last fully connected layer. Activation function used 
in convolutional and fully connected layers should avoid the problem of gradient loss. To cope with this 
problem, we use the following ReLU (Rectified Linear Unit) function \cite{Relu} as an activation function: 

\begin{eqnarray}
\centering {
f({x})=\max{(x,0)}\label{relu}
}
\end{eqnarray}

\begin{table*}[htb]
\caption{Details of the proposed CNN}
\label{cnn_table}
\begin{center}
\begin{tabular}{p{0.40\columnwidth}|p{0.20\columnwidth}p{0.20\columnwidth}p{0.30\columnwidth}p{0.30\columnwidth}p{0.30\columnwidth}}
\hline \hline
\hfil Layer&\hfil Filter&\hfil Function&\hfil Dropout&\hfil Pooling&\hfil Output size \\ \hline
\hfil Convolutional Layer 1A$\cdot$1B&\hfil 16$\times$16&\hfil ReLU&\hfil -&\hfil 2$\times$2&\hfil 55$\times$55$\times$32\\
\hfil Convolutional Layer 2A$\cdot$2B&\hfil 8$\times$8&\hfil ReLU&\hfil - &\hfil 2$\times$2&\hfil 24$\times$24$\times$64\\
\hfil Convolutional Layer 3&\hfil 5$\times$5&\hfil ReLU&\hfil - &\hfil 2$\times$2&\hfil 10$\times$10$\times$64\\
\hfil Convolutional Layer 4&\hfil 3$\times$3&\hfil ReLU&\hfil - &\hfil 2$\times$2&\hfil 4$\times$4$\times$64\\
\hfil Fully-connected Layer 1&\hfil -&\hfil ReLU&\hfil 0.5 &\hfil -&\hfil 1$\times$1$\times$1024\\
\hfil Fully-connected Layer 2&\hfil -&\hfil ReLU&\hfil 0.5 &\hfil -&\hfil 1$\times$1$\times$1024\\
\hfil Fully-connected Layer 3&\hfil -&\hfil softmax&\hfil - &\hfil -&\hfil 1$\times$1$\times$2\\ \hline \hline
\end{tabular} 
\end{center}
\end{table*}

\subsection{Discriminator}\label{sec:discriminator}

Our CNN predicts whether or not a robot can successfully pick an object. 
Given 250$\times$250 [pixel] subset of a pile's depth image and a pose of parallel jaw gripper, 
the CNN outputs the success rate of picking. 
If the success rate is more than 0.5, we judge that a robot will successfully pick an object. 
Otherwise, we judge that a robot will fail in picking an object. 

\subsection{Optimum Grasping Pose Detection}\label{Approach}

\begin{figure}[htbp]
\centering {
\includegraphics[width=90mm]{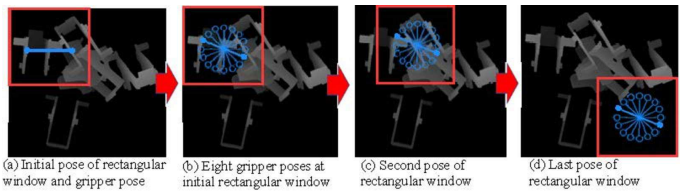}
}
\caption{Method for detecting graspable positions}
\label{estimation}
\end{figure}

To detect the optimum grasping pose, 
Lenz et al.\cite{Lenz} used a 2 step DNN (Deep Neural Network) where 
multiple candidates of grasping poses are generated by using a small Neural Network in the first step, 
and then optimal grasping pose is detected by using a larger Neural Network in the second step. 
However, this method requires high calculation cost since this method uses Raster scan with changing the 
size and orientation of a rectangular window and iteratively uses two DNNs. 
On the other hand, we apply a simple method to detect the grasping pose maximizing the success rate of picking 
(Fig. \ref{estimation}).  
Our method uses raster scan with fixed size and orientation of a rectangular window. 
We consider eight candidates of gripper's orientation corresponding to each rectangular window. 
By considering 6$\times$6 candidates of gripper's position, we totally have 
288 (8$\times$6$\times$6) candidates of gripper's grasping poses. 
For each gripper's pose, we calculate success rate of picking by using CNN. 
Among, 288 candidates, we consider 
calculating a grasping pose with highest success rate of picking. 

\begin{table}[htbp]
\begin{center}
\caption{Classification results of the verification data}
\begin{tabular}{cc|c|c}
\hline
& & \multicolumn{2}{c}{Simulation} \\
 & & ~~~~~Success~~~~~ & ~~~~Failure~~~~~\\ \hline \hline
 & Success & 436(TP)& 134(FP) \\
\cline{2-4}
\multirow{-2}{*}{~Discriminator~} & ~~Failure~~ & 164(FN)& 466(TN) \\ \hline
\end{tabular}
\\
\vspace{2mm}
\bf{Precision=0.765, Recall=0.727, F-value=0.745}
\label{judge_table}
\end{center}
\end{table}

\begin{figure}[thb]
\centering {
\includegraphics[width=65mm]{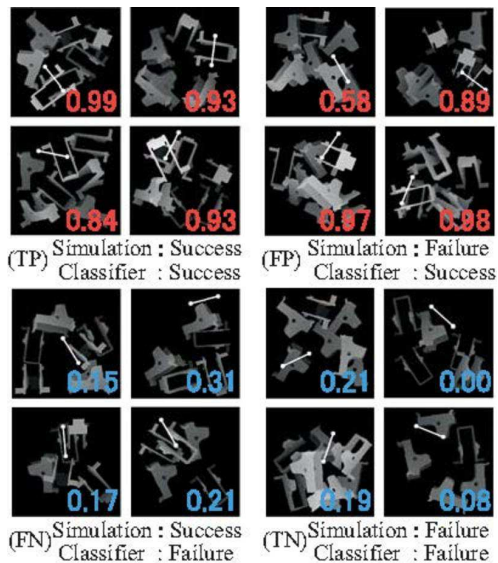}
}
\caption{Classification examples of the verification data. The numerical values indicate the estimated success rate by using the proposed
CNN. The values in red indicate a successful picking classification, while the values in blue indicate a failure picking classification.}
\label{judge_figure}
\end{figure}
%

\section{Collection of Training Data}\label{sec:training}

We performed physics simulation of bin-picking for 6 hours with 15 threads and collected 
6000 success data. The failure data is sampled to make the number of failure data be same as the number of success data. 
90$\%$ of the data is used to train the CNN and remaining 10 $\%$ is used to 
verify the trained CNN. 
By rotating and inverting the depth image included in the training data, 
we extend the number of training data up to 64800. 
By using the training data, we trained the CNN shown in Fig. \ref{cnn_figure} for 17 hours. 

\section{Results}\label{sec:results}
\subsection{Discrimination}\label{sec:discrimination}

As shown in Table \ref{judge_table}, 
we verified the trained CNN by using 1200 verification data including 600 success and 600 failure cases. 
Fig. \ref{judge_figure} shows 4 examples included in four classes 
(TP:True Positive),(TN:True Negative),(FP:False Positive) and (FN:False Negative) shown in Table \ref{judge_table} where 
red and blue figures respectively show the success and failure cases. 
We judged the successful cases if the success rate is larger than 0.5. 
F-value of our discriminator is 0.745 including the cases where a robot successfully picked up an object 
in spite of the prediction result where a robot fails in picking up an object. We will analyze 
this prediction error in more detail in the following subsections. 

%
\begin{figure*}[htb]
\centering {
\includegraphics[width=155mm]{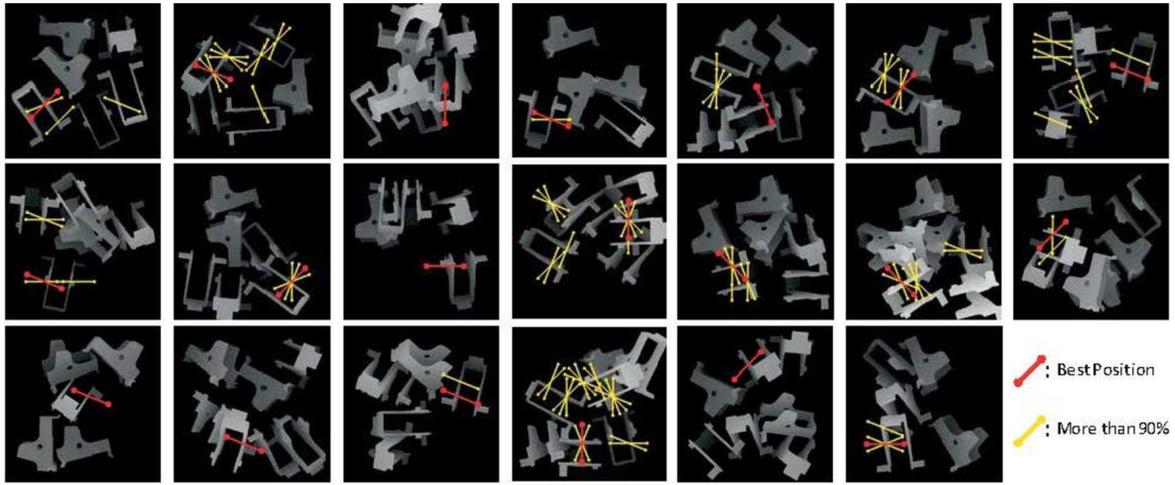}
}
\caption{Detection results of the graspable position by using the proposed method}
\label{fig:detection}
\end{figure*}
\subsection{Derivation of Optimum Grasping Pose}\label{最適把持位置の導出結果}

\begin{figure}[htb]
\centering {
\includegraphics[width=85mm]{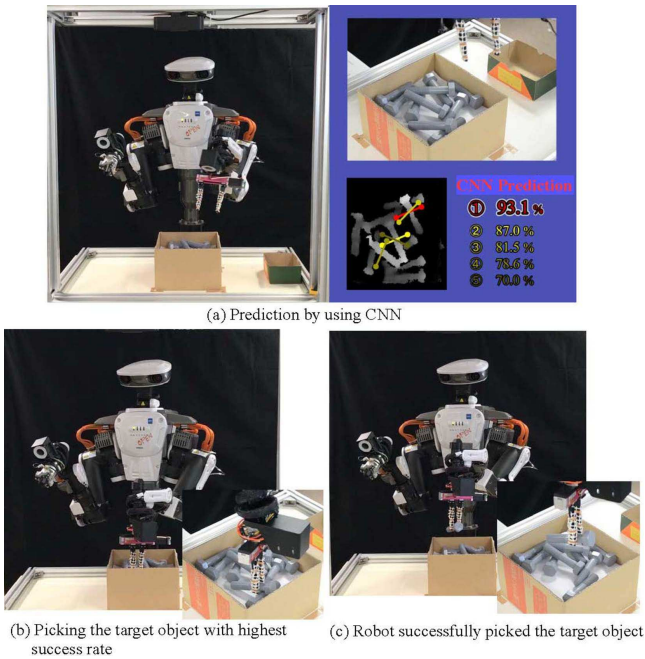}
}
\caption{Experimental result}
\label{experiment}
\end{figure}

By using the trained CNN and 20 verification data, we detected the optimum grasping pose as shown in Fig. \ref{fig:detection} where the segments marked in red and yellow shows the optimum grasping pose and grasping poses where the success rate is more 
than 0.9, respectively. 
We confirmed that, in all cases, the obtained grasping poses have high graspability index\cite{Domae2014}. 
Fig. \ref{experiment} shows an experimental result where, for given depth image of the pile, we determined 
the grasping pose by using CNN trained by using physics simulation.

\subsection{Analysis of Model Approximation}\label{sec:approximation}

Let us consider the effect of approximation introduced in our physics simulation. 
We consider fitting a rectangular to each convex decomposed part of a grasped object. 
While this approximation is used for checking collision among objects and fingers, 
a simulated depth image is obtained by using object models with the original shape. 
In our physics simulation, a robot sometimes stably grasps a part of an approximated shaped object 
where a robot may not be able to stably grasp the same part of an original shaped object. 
However, even if we use such unrealistic training data caused by the effect of approximation, 
the effect of approximation may be relaxed if we use the depth image of original shaped object 
as an input to CNN. 

\begin{figure}[thb]
\centering {
\includegraphics[width=75mm]{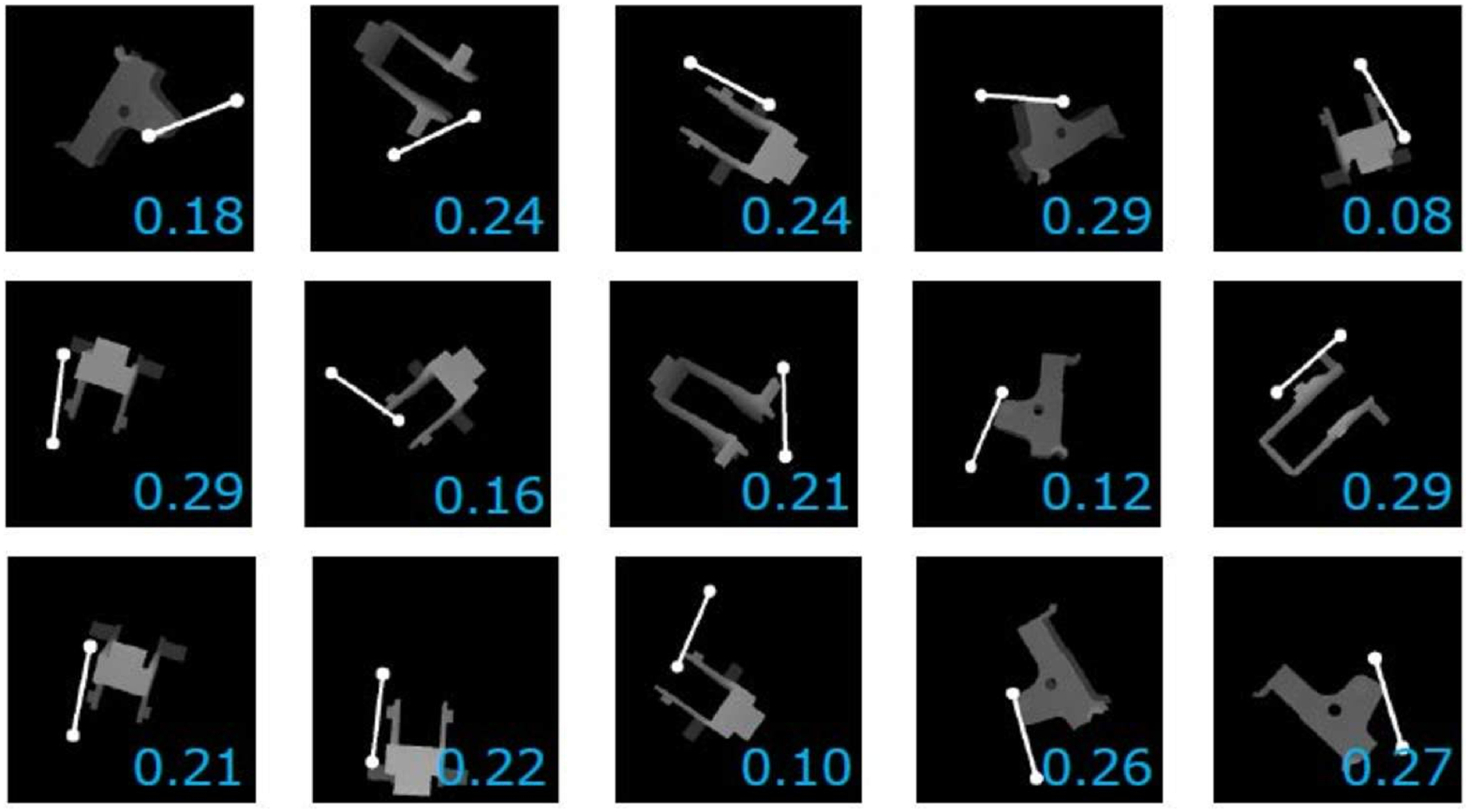}
}
\caption{Successful pickings in simulation due to the approximation. Blue values indicate estimated success rate by the proposed CNN.}
\label{fig:approximation}
\end{figure}

To explain this phenomenon, we collected 200 cases of physics simulation where 
a robot successfully picked an isolated single object. 
Among 200 cases, we picked up 15 unrealistic cases as shown in Fig. \ref{fig:approximation} 
where a robot stably grasps an object contrary to our expectations. 
In these cases, a robot stably grasps a part of an object with approximated shape while 
this part is not included in an object with original shape. 
The figure also shows the success rate obtained by using the trained CNN. 
The interesting thing is that the success rate is low in most of the cases in spite of the 
fact that a robot successfully picks an object in the physics simulation. 
This implies that, in the discrimination result shown in Table \ref{judge_table}, 
(FP) and (FN) do not simply show the cases of discrimination errors. 
We can consider that the effect of approximation is relaxed if we use 
the depth image including original shaped objects to train the CNN. 

\begin{figure}[thb]
\centering {
\includegraphics[width=85mm]{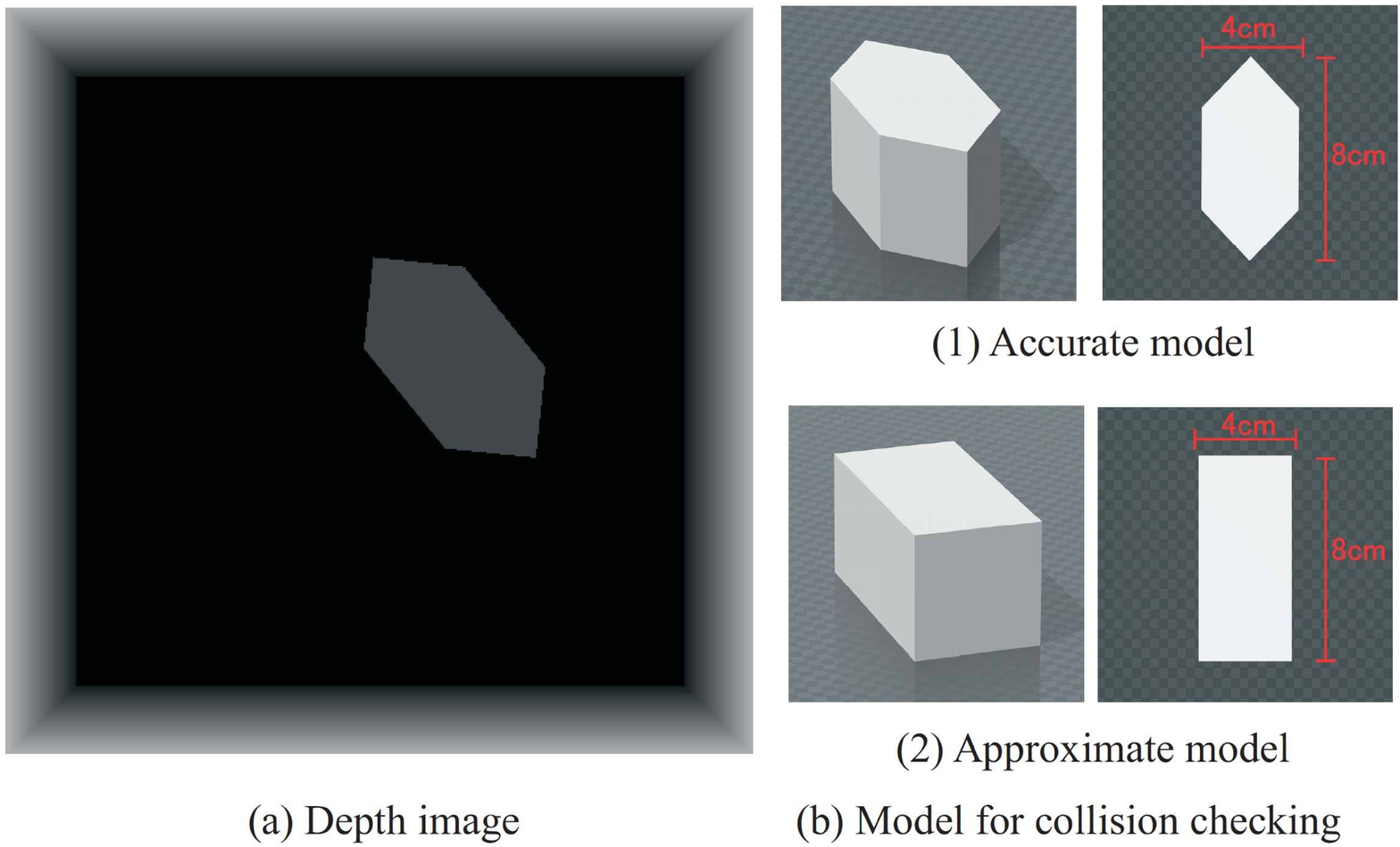}
}
\caption{Successful pickings in simulation due to the approximation. Blue values indicate estimated success rate by the proposed CNN.}
\label{verifySimulation}
\end{figure}

\begin{figure}[thb]
\centering {
\includegraphics[width=85mm]{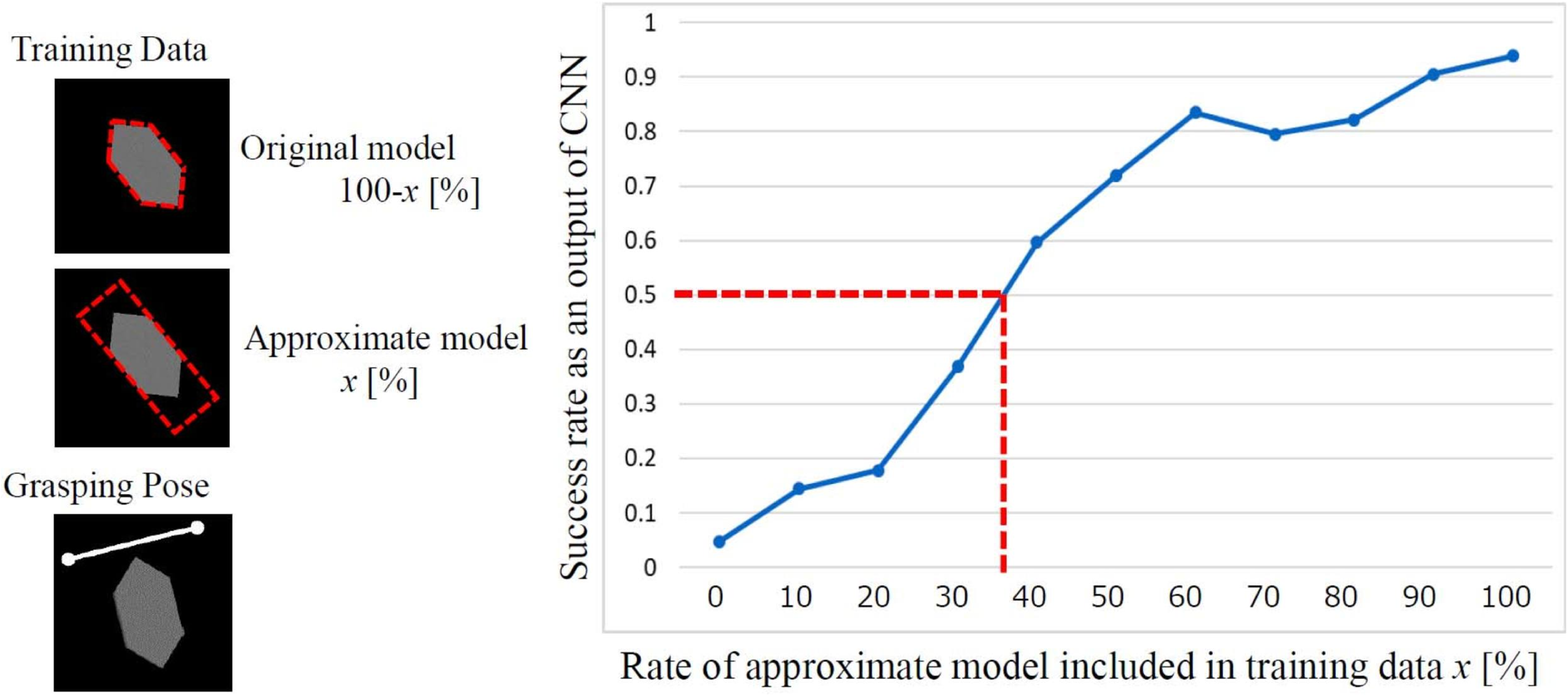}
}
\caption{Successful pickings in simulation due to the approximation. Blue values indicate estimated success rate by the proposed CNN.}
\label{obj1Simulation}
\end{figure}

\begin{figure}[thb]
\centering {
\includegraphics[width=85mm]{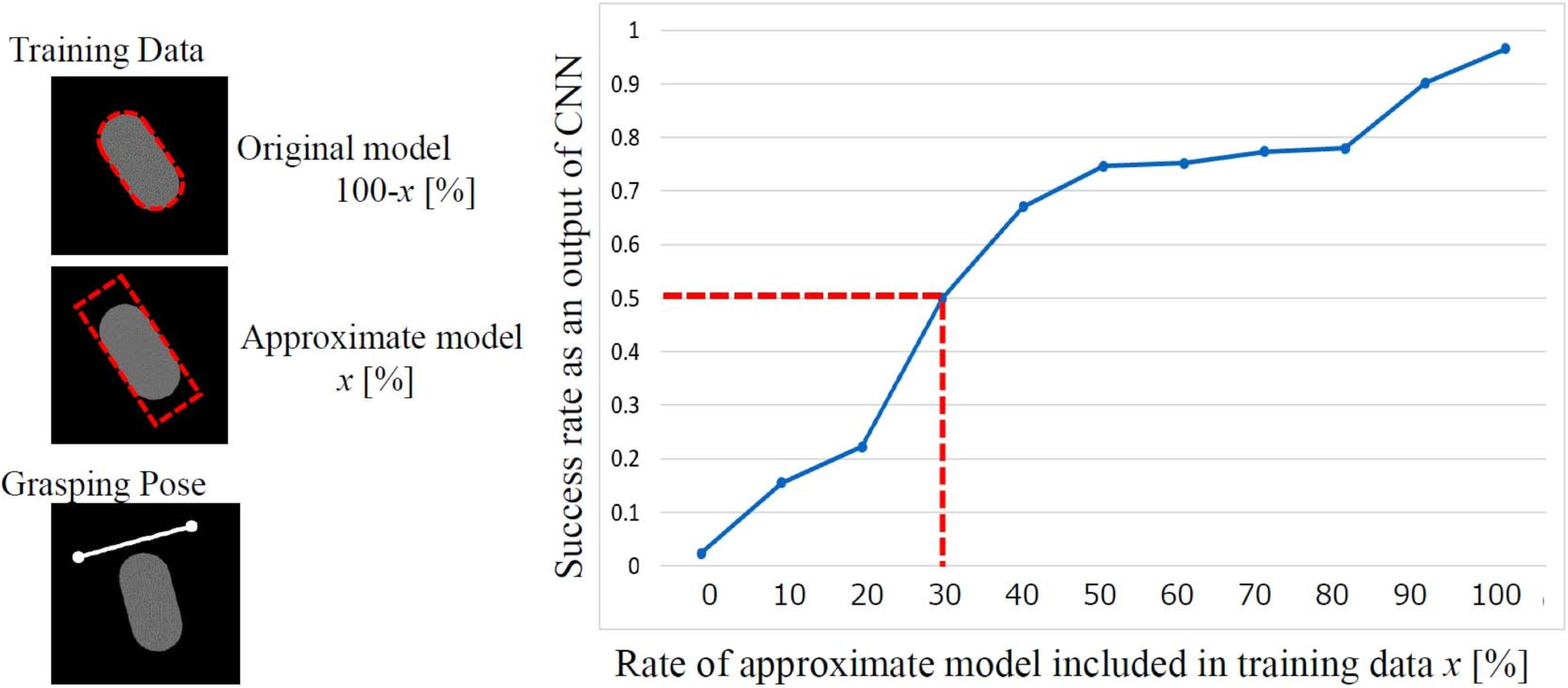}
}
\caption{Successful pickings in simulation due to the approximation. Blue values indicate estimated success rate by the proposed CNN.}
\label{obj2Simulation}
\end{figure}

Let us consider analyzing the effect of shape approximation in more detail. As shown in 
Fig. \ref{verifySimulation}, we consider making a robot pick an object placed on a tray. 
We prepared two kinds of objects to train the CNN where one is approximated by rectangular parallelepiped 
and the other is not approximated when checking collision. 
We change the rate of using approximated object when training the CNN. After finished training the CNN, 
we consider estimating the success rate 
when a robot trying to pick a part of an object where it is included 
approximated one and is not included in the original one. 
The results are shown in Figs. \ref{obj1Simulation} and \ref{obj2Simulation} where 
a hexagonal prism and an elliptic cylinder are used, respectively. 
In both cases, if the rate of using approximated object is less than 30 $\%$, 
we can correctly estimate the success of the picking since success is predicted if the success rate is 
larger than 0.5. This result means that, in randomized bin-picking, we can correctly estimate 
whether or not a robot can successfully pick an object from the pile if such rough approximation is 
used in less than 30 $\%$ of rectangular. 

%
%


\section{Conclusions}\label{sec:conclusion}

In this research, we researched the learning based randomized bin-picking. 
We introduced approximate physics simulation to effectively collect the training data within 
short period of time. We first formulated the learning based method by using CNN. Then, we obtained 
the optimum grasping posture of parallel jaw gripper by using CNN. 
Finally, we showed that the effect of approximation introduced in collision checking is relaxed 
if we use exact 3D model to generate the depth image of the pile as an input to CNN.

\end{document}